\documentclass[runningheads]{llncs}

\usepackage[mobile]{eccv}

\usepackage{eccvabbrv}

\usepackage{wrapfig}

\usepackage{graphicx}
\usepackage{booktabs}
\usepackage{arydshln}
\usepackage{verbatim}
\usepackage[accsupp]{axessibility}  %

\usepackage{hyperref}

\usepackage{orcidlink}
\usepackage{graphicx}

\usepackage[parfill]{parskip}
\usepackage{amsmath}
\usepackage{algorithm}
\usepackage{algpseudocode}

\usepackage{float}
\usepackage{multirow}
\usepackage{soul}

\definecolor{DarkRed}{rgb}{0.8, 0.0, 0.4}
\definecolor{DarkGreen}{rgb}{0.0, 0.6, 0.4}
\definecolor{MetaBlue}{rgb}{0.0, 0.51, 0.98}

\newcommand{\methodname}{Imagine Flash}
\newcommand{\teacher}{\Phi}
\newcommand{\student}{\Theta}
\newcommand{\lossname}{SRL}
\newcommand{\x}{\mathbf{x}}
\newcommand{\eps}{\boldsymbol{\epsilon}}
\newcommand{\figlabel}{Fig.~}

\begin{document}

\title{\methodname: Accelerating Emu Diffusion Models with Backward Distillation}

\def\thefootnote{*}\footnotetext{Equal Contribution}\def\thefootnote{\arabic{footnote}}

\titlerunning{\methodname: Accelerating Emu with Backward Distillation}

\newcommand\blfootnote[1]{%
  \begingroup
  \renewcommand\thefootnote{}\footnote{#1}%
  \addtocounter{footnote}{-1}%
  \endgroup
}

\author{Jonas Kohler$^*$ \and
Albert Pumarola$^*$ \and
Edgar Sch{\"o}nfeld$^*$\and
Artsiom Sanakoyeu$^*$ \and
Roshan Sumbaly \and
Peter Vajda \and
Ali Thabet
}

\authorrunning{ }

\institute{GenAI, Meta\\
\email{\{jonaskohler,apumarola,edgarschoenfeld,asanakoy\}@meta.com}
}

\maketitle

\begin{figure}[h]
  \centering
  \vspace{-0.8cm}
  \makebox[\textwidth]{\includegraphics[width=1.1975\linewidth]{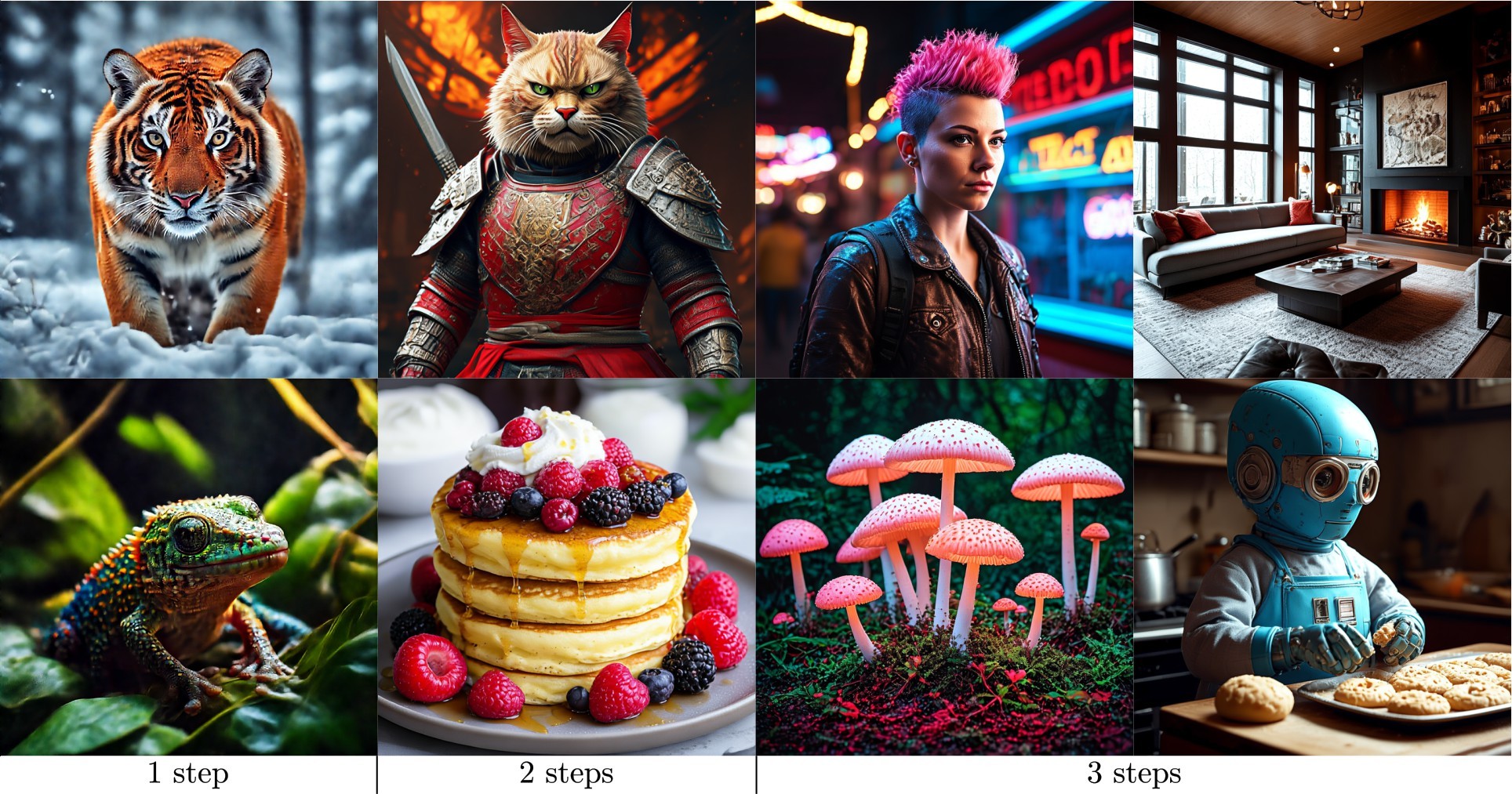}}
\vspace{-0.4cm}
  \caption{
  \footnotesize
  \textbf{\methodname\ generation in 1, 2, and 3 steps}. \methodname\ uses backward distillation to accelerate inference of a baseline diffusion model (Emu). Our distillation framework allows generating high quality images with as few as 1-3 steps. %
  }
  \vspace{-0.7cm}
  \label{fig:teaser}
\end{figure}

\vspace{-7.5pt}
\begin{abstract}
Diffusion models are a powerful generative framework, but come with expensive inference.
Existing acceleration methods often compromise image quality or fail under complex conditioning when operating in an extremely low-step regime. In this work, we propose a novel distillation framework tailored to enable high-fidelity, diverse sample generation using just one to three steps. Our approach comprises three key components: (i) Backward Distillation, which mitigates training-inference discrepancies by calibrating the student on its own backward trajectory; (ii) Shifted Reconstruction Loss that dynamically adapts knowledge transfer based on the current time step; and (iii) Noise Correction, an inference-time technique that enhances sample quality by addressing singularities in noise prediction. Through extensive experiments, we demonstrate that our method outperforms existing competitors in quantitative metrics and human evaluations.
Remarkably, it achieves performance comparable to the teacher model using only three denoising steps, enabling efficient high-quality generation.
\end{abstract}

\section{Introduction}
\label{sec:intro}
Generative modelling has witnessed a paradigm shift with the advent of Denoising Diffusion Models (DMs) \cite{ho2020denoising,song2020score}. These models have set new benchmarks across various domains \cite{rombach2022high,kong2020diffwave,ho2022imagen}, offering an unprecedented combination of realism and diversity, while ensuring stable training. However, the sequential nature of the denoising process presents a significant challenge. Sampling from DMs is a time-consuming and costly process, with the time required largely dependent on two factors: (i) the latency of the per-step neural network evaluation, and (ii) the total number of denoising steps.

Considerable research efforts have been devoted to accelerating the sampling process. For text to image synthesis, the proposed methods span a wide range of techniques, including higher-order solvers \cite{lu2022dpm}, modified diffusion formulations for curvature reduction \cite{liu2022flow}, as well as guidance- \cite{meng2023distillation}, step-\cite{salimans2022progressive} and consistency distillation \cite{song2023consistency}. These methods have brought impressive improvements, achieving very high quality in the close to 10 step regime. More recently, hybrid methods that leverage both distillation and adversarial losses \cite{sauer2023adversarial, xu2023ufogen, lin2024sdxl} have pushed the boundary to under five steps. While these methods achieve impressive quality on simple prompts and uncomplicated styles like animation, they suffer from degraded sample quality for photorealistic images, especially for complex text conditioning. 

A common theme among the aforementioned methods is the attempt to align the few-step student model with the complex teacher paths, despite endowing the student model with significantly lower capacity (i.e., steps). Recognizing this as a limitation, we invert the process by proposing a novel distillation framework that is designed for the teacher to improve the student along it's own diffusion paths. In summary, our contribution is threefold:

\begin{itemize}
\item First, our approach introduces Backward Distillation, a distillation process designed to calibrate the student model on its own upstream backward trajectory, thereby reducing the gap between the training and inference distributions and ensuring zero data leakage during training across all time steps.

\item Second, we propose a Shifted Reconstruction Loss that dynamically adapts the knowledge transfer from the teacher model. %
Specifically, the loss is designed to distill global, structural information from the teacher at high time steps, while focusing on rendering fine-grained details and high-frequency components at lower time steps. This adaptive approach enables the student to effectively emulate the teacher's generation process at different stages of the diffusion trajectory.%

\item Finally, we propose Noise Correction, an inference-time modification that enhances sample quality by addressing singularities present in noise prediction models during the initial sampling step. This training-free technique mitigates degradation of contrast and color intensity that usually arise when operating with an extremely low number of denoising steps.
\end{itemize}

\begin{figure}[t]
\centering
\includegraphics[width=\linewidth]{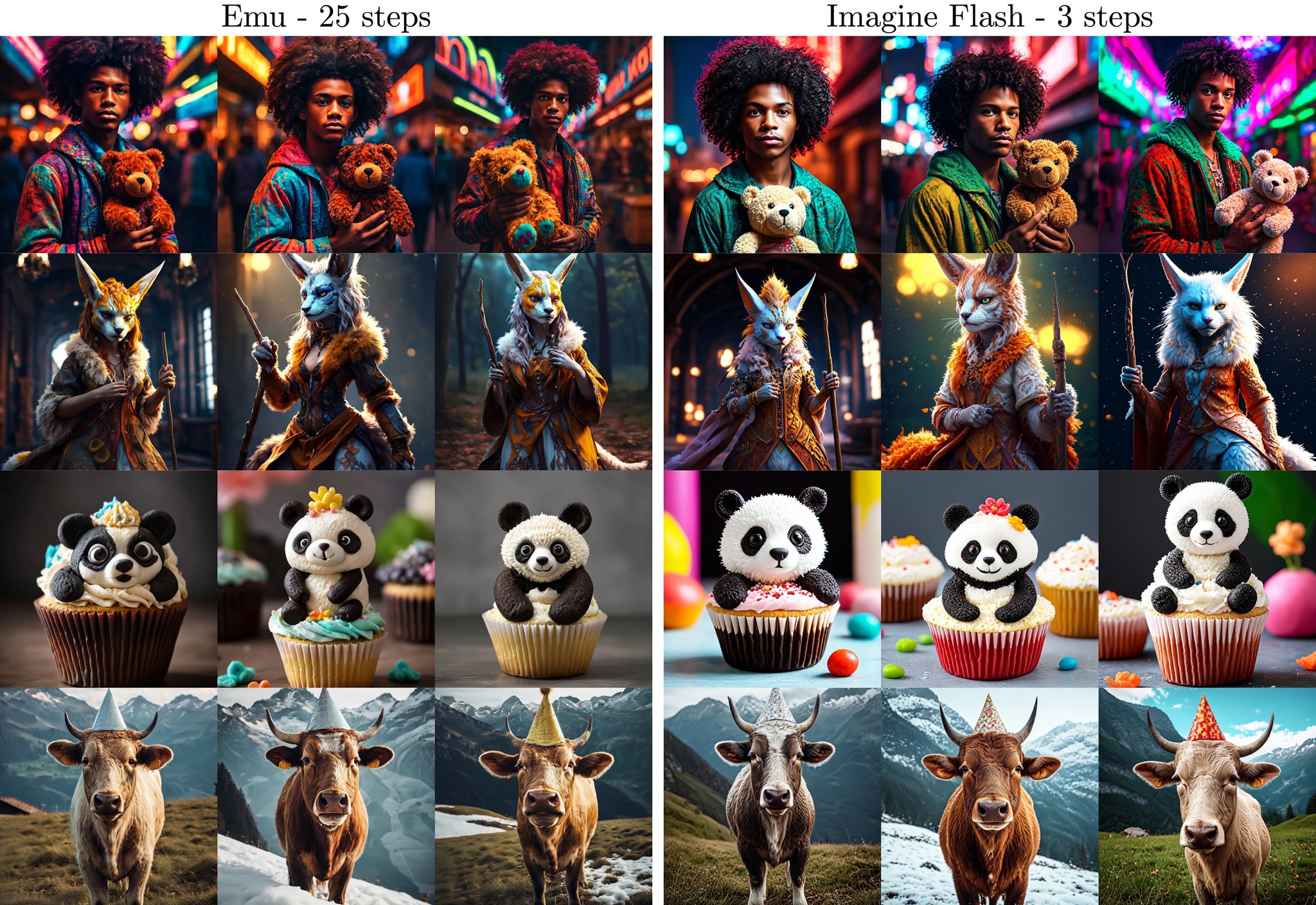}
  \vspace{-0.6cm}
  \caption{
  \footnotesize 
  \textbf{\methodname~vs Emu}. Our method reduces the inference time of the baseline by a significant amount, while still generating high quality and complex images. 
  }
  \vspace{-0.1cm}
  \label{fig:intro}
\end{figure}

By synergistically combining these three novel components, we apply our distillation framework to a baseline diffusion model, Emu~\cite{dai2023emu}, resulting in \methodname\, which achieves high-quality generation in the extremely low-step regime without compromising sample quality or conditioning fidelity (\figlabel \ref{fig:intro}). Through extensive experiments and human evaluations, we demonstrate the effectiveness of our approach in achieving favorable trade-offs between sampling efficiency and generation quality across a range of tasks and modalities.

\section{Related Work}
\label{sec:related_work}

Diffusion models \cite{sohl2015deep,ho2020denoising,nichol2021improved}, in contrast to previous generative models (\eg, GANs), approach density estimation and data sampling in an iterative way, by gradually reversing a noising process. This iterative nature translates to multiple queries of a neural network backbone, leading to high inference costs. As a result, a large body of works has focused on producing faster and more efficient ways of sampling from diffusion models. However, enhancing inference speed without sacrificing image quality and text faithfulness continues to present a considerable challenge.

\textbf{Solvers and curvature rectification:} Early approaches focus on developing better solvers to the underlining dynamics of the diffusion process. Along this line, several works propose exponential integrators \cite{lu2022dpmplus,zhang2022fast}, higher order solvers \cite{zhao2023unipc, karras2022elucidating} and model-specific bespoke solvers \cite{shaul2023bespoke,zheng2023dpm}. Other studies investigate reformulations of the diffusion process with the aim of minimizing curvature in both the forward (noising) \cite{albergo2023stochastic,lipman2022flow} and backward (de-noising) trajectories \cite{liu2022flow,pooladian2023multisample,lee2023minimizing, karras2022elucidating}. In a nutshell, these approaches aim to linearize the inference path, allowing for larger step sizes, and therefore fewer steps at inference time. Despite the substantial step reduction of these methods, there is a limit on how large the inference step can be, without compromising image quality. 

\textbf{Reducing model size:} A series of orthogonal works looks at reducing the per-step cost. In this vein, several works focus on employing smaller backbone architectures \cite{peebles2023scalable,yang2023diffusion,li2023snapfusion}, and even mobile friendly networks \cite{zhao2023mobilediffusion}. Reducing per-step cost is also addressed by minimizing the cost of conditional generation at each iteration or caching intermediate activations in the network backbone \cite{wimbauer2023cache}. To that extent,  
\cite{meng2023distillation} propose guidance distillation, while \cite{castillo2023adaptive} present the training-free alternative of truncating guidance. Reducing per-step latency leads to significant gains in inference speed. However, to truly scale inference for real-time applications, these advancements must be coupled with further reductions of the number of steps to a small single-digit number.

\textbf{Reducing sampling steps:} One way to further reduce inference latency is step distillation \cite{salimans2022progressive,meng2023distillation}. In these work, the authors propose a progressive approach to distill two or more steps into a single one. The effect is further enhanced when using consistency constraints \cite{luo2023latent,song2023consistency}. While these approaches achieve significant step reductions, substantial quality degradation is evident at low step regimes. To compensate for quality loss, a further line of work proposes additional training enhancements during distillation. Namely, ADD~\cite{sauer2023adversarial}, Lightning~\cite{lin2024sdxl} and UFOGEN~\cite{xu2023ufogen} add an adversarial loss to increase sample quality. 

While the above distillation methods undoubtedly produce impressive results with just a single step generation, these improvements are still not adequate for many practical applications, such as generating photorealistic images or adhering to complex prompts.
A more reasonable approach is to control the trade-off between quality and speed. Practically, this translates to methods that allow small step increases (2 to 4 steps), with a major gain in quality. We adopt this approach in our method. To achieve better quality, we propose to distill along the student's \textit{backward} path instead of the forward path. Put differently, rather than having the student mimic the teacher, we use the teacher to improve the student based on its current state of knowledge. We find that this approach leads to competitive results with single inference steps, and substantially improves quality and fidelity with just a slight increase to as few as 3 steps.

\section{Background on Diffusion Models}
Diffusion models consist of two interconnected processes: forward and backward. The forward diffusion process gradually corrupts the data by interpolating between a sampled data point $\mathbf{x}_0$ and Gaussian noise $\boldsymbol{\epsilon} \sim \mathcal{N}(0,\textbf{I})$. 
That is 
\begin{equation}\label{eq:forward}
\mathbf{x}_t = q(\mathbf{x}_0, \boldsymbol{\epsilon}, t) = \alpha_t \mathbf{x}_0 + \sigma_t \boldsymbol{\epsilon},\quad \forall t\in [0,T],
\end{equation}
where $\alpha_t$ and $\sigma_t$ define the signal-to-noise ratio (SNR) of the stochastic interpolant $\mathbf{x}_t$. In the following, we opt for coefficients $(\alpha_t,\sigma_t)$ that result in a variance-preserving process (see e.g. \cite{karras2022elucidating}). When viewed in the continuous time limit, the forward process in Eq.~\ref{eq:forward} can be expressed as the stochastic differential equation (SDE) $d\mathbf{x} = \mathbf{f}(\mathbf{x},t)dt + g(t)d\mathbf{w_t}$, where $f(\mathbf{x},t): \mathbb{R}^d\rightarrow \mathbb{R}^d$ is a vector-valued drift coefficient, $g(t):\mathbb{R}\rightarrow\mathbb{R}$ is the diffusion coefficient, and $\mathbf{\mathbf{w}_t}$ denotes the Brownian motion at time $t$.

Inversely, the backward diffusion process is intended to undo the noising process and generate samples. 
According to Anderson's theorem \cite{anderson1982reverse}, the forward SDE introduced earlier satisfies a reverse-time diffusion equation, which can be reformulated using the Fokker-Planck equations \cite{song2020score} to have a deterministic counterpart with equivalent marginal probability densities, 
known as the \textit{probability flow ODE}:
\begin{equation}\label{eq:pflowode}
d\mathbf{x} = \left[\mathbf{f}(\mathbf{x},t) - \frac{1}{2} g(t)^2 \nabla_{\mathbf{x}} \log p_t(\mathbf{x})\right]dt.
\end{equation}
As demonstrated in~\cite{hyvarinen2005estimation,song2020score}, this marginal transport map can be learned through maximum likelihood estimation of the perturbation kernel of diffused data samples $\nabla_\mathbf{x} \log p_t(\mathbf{x}|\mathbf{x}_0)$ in a simulation-free manner. This allows us to estimate $ \hat{\boldsymbol{\epsilon}}(\mathbf{x}_t,t) / \sigma_t \approx \nabla_\mathbf{x} \log p_t(\mathbf{x}|\mathbf{x}_0) $,
usually parameterized by a time-conditioned neural network. Given these estimates, we can sample using an iterative numerical solver $f$~\cite{song2020denoising}:
\begin{equation}\label{eq:backward_iter}
\mathbf{x}_0 \approx f \circ f \circ \cdots \circ f (\mathbf{x}_T).
\end{equation}
without loss of generality, in the paper we use the update rule given by first-order solvers like DDIM\cite{song2020denoising}, i.e.:
\begin{equation}\label{eq:backward}
\mathbf{x}_{t-1} = f(\mathbf{x}_t) = \alpha_{t-1} \hat{\mathbf{x}}_0(\mathbf{x}_t, \hat{\boldsymbol{\epsilon}}, t) + \sigma_{t-1}\hat{\boldsymbol{\epsilon}}(\mathbf{x}_t,t),
\end{equation}
where the sample data estimate $\hat{\mathbf{x}}_0$ at time-step $t$ is computed as:
\begin{equation}\label{eq:x0_hat}
\hat{\mathbf{x}}_0(\mathbf{x}_t,\hat{\boldsymbol{\epsilon}},t) = \frac{\mathbf{x}_t - \sigma_t\hat{\boldsymbol{\epsilon}}(\mathbf{x}_t,t)}{\alpha_t}.
\end{equation}

\section{Method}
We present \methodname, a novel distillation technique designed for fast text-to-image generation that builds upon -- but is not exclusive to --  Emu~\cite{dai2023emu}. Unlike the original Emu model that requires at least $50$ neural function evaluations (NFEs) to produce high-quality samples, \methodname\ achieves comparable results with just a few NFEs. Our proposed distillation method comprises three novel key components: (i) \textit{Backward Distillation}, a distillation process that ensures zero data leakage during training for all time points $t$ (see Sec.~\ref{sec:backward_distillation}). (ii) \textit{Shifted Reconstruction Loss (\lossname)}, an adaptive loss function designed to maximize knowledge transfer from the teacher (see Sec.~\ref{sec:strucutre_guidance}). (iii) \textit{Noise Correction}, a training-free inference modification that improves the sample quality of few-step methods that were trained in noise prediction mode (see Sec.~\ref{sec:noise_correction}).

In what follows, we assume access to a pre-trained diffusion model $\teacher$, which predicts noise estimates $\hat{\boldsymbol{\epsilon}}_\teacher$. This \textit{teacher model} can operate in either image or latent space. Our goal is to distill the knowledge of $\teacher$ into a \textit{student model} $\student$, while reducing the overall number of sampling steps, and providing high quality increases per extra step allowed in $\student$. If the $\teacher$ model uses classifier-free guidance (\textsc{cfg}), then we also distill this knowledge into our student and eliminate the need for \textsc{cfg}.

\subsection{Backward Distillation}
\label{sec:backward_distillation}

\begin{figure}[t]
\centering
\includegraphics[width=\linewidth]{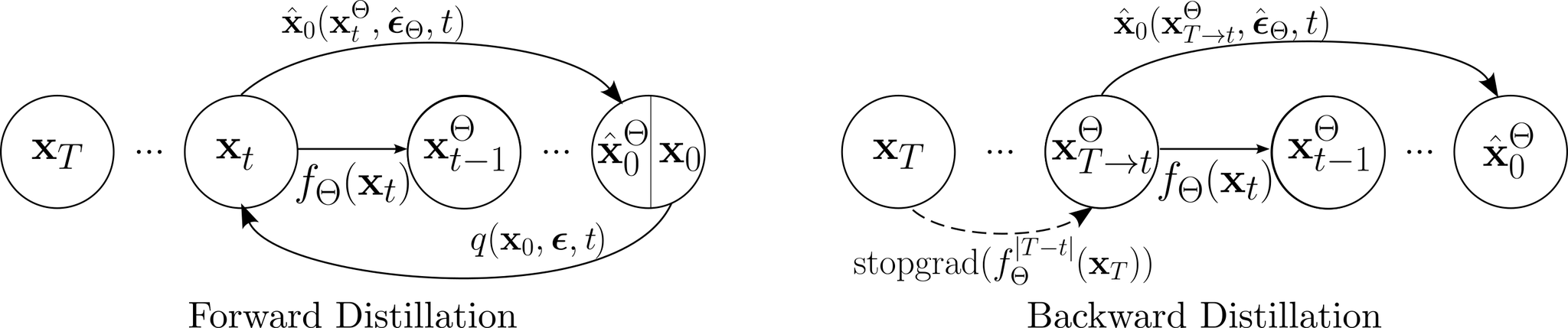}
  \caption{
  \footnotesize 
  \textbf{Backward Distillation} ensures signal consistency between training and inference during distillation. In standard forward distillation we run training steps starting from the forward-noised latent code $\mathbf{x}_t$. In backward distillation, we instead use the student model’s backward iterations to get latent $\mathbf{x}_{T \rightarrow t}^\student$ and use it as the starting code for training steps, where we compute gradients. Backward distillation eliminates information leakage for all $t$s, and prevents the model from relying on GT signals.
  }
  \label{fig:backward_distillation}
\end{figure}

It is widely recognized that conventional noise schedulers often fail to achieve zero terminal SNR at $t=T$~\cite{lin2024common}, thereby creating a discrepancy between training and inference. Specifically, the noise schedule ($\alpha_T, \sigma_T$) in Eq.~\ref{eq:forward} is commonly chosen s.t. $\mathbf{x}_T$ is not pure noise during training, but rather contains low-frequency information leaked from $\mathbf{x}_0$. This discrepancy leads to performance degradation during inference, especially when taking only a few steps. To overcome this issue, Lin~\textit{et al.}~\cite{lin2024common} suggest to rescale existing noise schedules under a variance-preserving formulation to enforce zero terminal SNR.

However, we argue that this solution is not sufficient as information leakage occurs not only at $t=T$, but at all $t$'s via the forward diffusion Eq.~\ref{eq:forward}. Recall that the distillation loss gradient is computed at every training step as follows:

\begin{equation}\label{eq:original_grad}
\nabla_\Theta \left\| \mathbf{x}_{t \rightarrow 0}^\teacher-\hat{\mathbf{x}}_0(\mathbf{x}_t,\hat{\boldsymbol{\epsilon}}_\student,t)  \right\|^2,
\end{equation}

where $\hat{\mathbf{x}}_0(\cdot)$ is defined in Eq.~\ref{eq:x0_hat} and 
$\mathbf{x}_{t \rightarrow 0}^\teacher=f^{|k|}_\teacher(\mathbf{x}_t)$ is the $k$ step teacher prediction. Now, since $\mathbf{x}_t=\alpha_t \mathbf{x}_0 + \sigma_t \mathbf{x}_T$ even when enforcing zero \textit{terminal} SNR ($\mathbf{x}_T=\boldsymbol{\epsilon}$) as suggested by \cite{lin2024common}, any stochastic interpolant $\mathbf{x}_t, t<T$ still contains information from the ground-truth sample via the first summand $\alpha_t\mathbf{x}_0$. As a result, the model learns to denoise taking into account information from the ground-truth signal. The smaller the $t$, the stronger the presence of the signal, and thus the more it will learn to preserve it. Let  $\mathbf{x}_{T \rightarrow t}^\student=f^{|T-t|}_\student(\mathbf{x}_T)$ be the student's estimate at time $t$ starting from pure noise at $T$ in $|T-t|$ steps (see Eq.~\ref{eq:backward}). During inference, the signal contained in $\mathbf{x}_{T \rightarrow t}^\student$ is \textit{no longer ground-truth} signal $\mathbf{x}_0$, but rather the student's own best guess $\mathbf{x}_0^{\student}:=\hat{\mathbf{x}}_0(\mathbf{x}_{t+1},\hat{\boldsymbol{\epsilon}}_\student,t+1)$ from the previous step (see Eq.~\ref{eq:backward_iter}). As a result, models that have been trained to preserve a given signal will continue to propagate errors from previous steps instead of correcting them.

We propose a solution to ensure signal consistency between training and inference at all times. This is achieved by simulating the inference process during training, a method we term \textit{backward distillation}. Unlike standard forward distillation, during training we do not begin sampling from the forward-noised latent code $\mathbf{x}_t=q(\mathbf{x}_0, \boldsymbol{\epsilon}, t)$. Instead, we first perform backward iterations of the student model to obtain $\mathbf{x}_{T \rightarrow t}^\student=\text{stopgrad}(f^{|T-t|}_\student(\mathbf{x}_T))$, and then use this as input for both the student and teacher models during training (see Fig.~\ref{fig:backward_distillation}). The training gradients are then computed as

\begin{equation}\label{eq:new_grad}
\nabla_\student \left\| \hat{\mathbf{x}}_0(f_\teacher^k(\mathbf{x}_{T \rightarrow t}^\student),\hat{\boldsymbol{\epsilon}}_\teacher, t/k)-\hat{\mathbf{x}}_0(\mathbf{x}_{T \rightarrow t}^\student,\hat{\boldsymbol{\epsilon}}_\student,t)  \right\|^2,
\end{equation}

where $\mathbf{x}_0^{\teacher}:=\hat{\mathbf{x}}_0(f_\teacher^k(\mathbf{x}_{T \rightarrow t}^\student),\hat{\boldsymbol{\epsilon}}_\teacher, t/k)$ represents the target produced by running the teacher for $k$ time-uniform denoising steps (from timestep $t$ to $t/k$) with \textsc{cfg}, starting from the current latent code $\mathbf{x}_{T \rightarrow t}^\student$.

In summary, \textit{backward distillation} eliminates information leakage at all time steps $t$, preventing the model from relying on a ground-truth signal. This is achieved by simulating the inference process during training, which can also be interpreted as calibrating the student on its own upstream backward path.\footnote{Note that the computation of $\mathbf{x}_{T \rightarrow t}^\student$ is cheap, as we distill for few steps only and gradient computation are omitted $\forall t'>t$.}

\textcolor{white}{EMU}
\subsection{{\lossname}: Shifted Reconstruction Loss}
~\label{sec:strucutre_guidance}

\begin{figure}[t]
\centering\includegraphics[width=\linewidth]{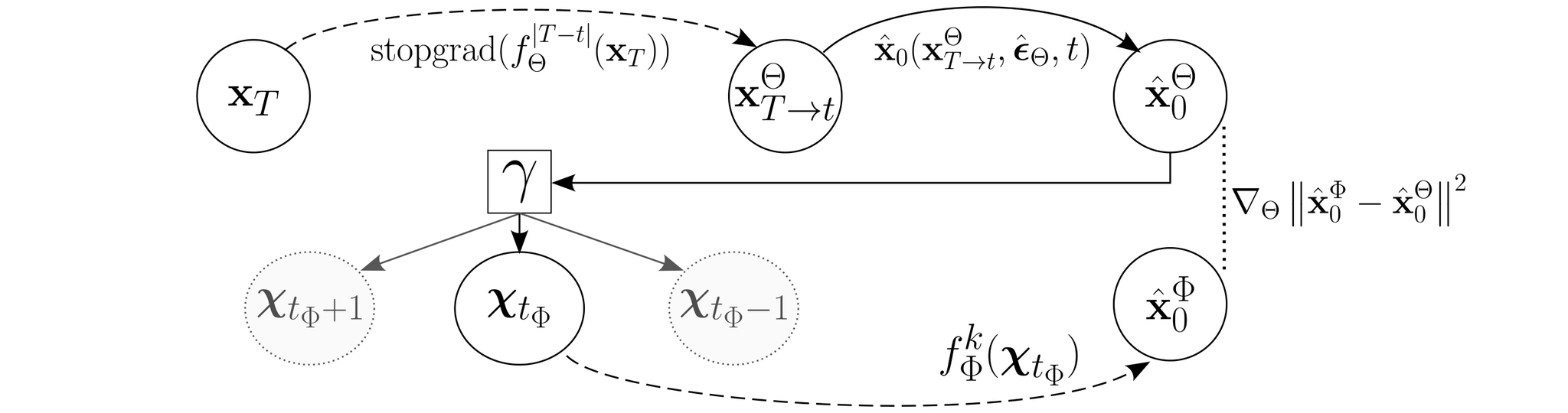}
  \caption{\footnotesize \textbf{Shifted Reconstruction Loss (\lossname)}. 
We propose a new distillation loss to improve the structure and adherence to detail of the student's predictions. $\mathbf{x}_{T \rightarrow t}^\student$ is the current noisy latent code at timestep $t$ in the context of backward distillation. $\hat{\mathbf{x}}_0^\student$ is the image predicted by the student in one step from $\mathbf{x}_{T \rightarrow t}^\student$. SRL then entails noising $\hat{\mathbf{x}}_0^\student$ again to a $t_\teacher$ specified by the shifting function $\gamma$, followed by $k$ uniformly sampled denoising steps from the teacher. The shifts are designed to adapt the type of knowledge distilled from the teacher for different t's in order to maximize efficacy.
}
  
  \label{fig:loss}
\end{figure}

In the process of image generation through backward diffusion, the early stages (where $t$ is close to $T$) are instrumental in formulating the overall structure and composition of the image. Conversely, the later stages (where $t$ is near $0$) are essential for adding high-level details~\cite{castillo2023adaptive}.
Drawing from this observation, we devise enhancements to the default knowledge distillation loss (Eq.~\ref{eq:original_grad}), that encourage the student model to learn both the structural composition and detail-rendering capabilities of the teacher model. This involves shifting starting points for the teacher denoising away from the student's  starting point $t$, hence we refer to this method as shifted reconstruction loss (\lossname). \figlabel \ref{fig:loss} provides an overview of our proposed loss. 

To obtain a target in \lossname, instead of running the teacher model from the current noisy latent code $\mathbf{x}_{T \rightarrow t}^\student$ as in Eq.~\ref{eq:new_grad}, we generate the target from the student's prediction  $\mathbf{x}_0^{\student}=\hat{\mathbf{x}}_0(\mathbf{x}_{T \rightarrow t}^\student,\hat{\boldsymbol{\epsilon}}_\student,t)$ noised to $t_\teacher=\gamma(t)$, which we term $\boldsymbol{\chi}_{t_\teacher}=q(\mathbf{x}_0, \eps, \gamma(t))$. As a result, the gradient updates are computed  as

\begin{equation}\label{eq:distill}
\nabla_\student \left\| \hat{\mathbf{x}}_0(f_\teacher^k(\boldsymbol{\chi}_{t_\teacher}),\hat{\boldsymbol{\epsilon}}_\teacher, \gamma(t)/k)-\hat{\mathbf{x}}_0(\mathbf{x}_{T \rightarrow t}^\student,\hat{\boldsymbol{\epsilon}}_\student,t)  \right\|^2.
\end{equation}

Unlike conventional step distillation methods \cite{meng2023distillation} where both the teacher and student begin with the same latent code, in SRL the mapping function $\gamma: [0,T] \rightarrow [0,T]$ is not defined as the identity function $\gamma(t):=t$. 
Instead, it is designed such that for higher values of $t$, the target produced by the teacher model displays global content similarity with the student output but with improved semantic text alignment; and for lower values of $t$, the target image features enhanced fine-grained details while maintaining the same overall structure as the student's prediction. This approach encourages the student to prioritize distilling structural knowledge during the early backward steps and focus on generating more refined details towards the final backward steps.

\subsection{Noise Correction}
\label{sec:noise_correction}
The most common diffusion models are trained in noise prediction mode \cite{ho2020denoising,rombach2022high}, which, according to Eq.~\ref{eq:forward}, tasks the network with separating noise from signal given a randomly corrupted image. However, the process of sampling from diffusion model naturally starts from a point of pure noise, i.e. $\x_T=\eps$, where $\eps\sim\mathcal{N}(0,\mathbf{I})$. Consequently, there is no signal to be found in $\mathbf{x}_T$, and hence noise prediction at $T$ becomes trivial but completely uninformative for the image generation process. To circumvent this singularity, existing works modify the noise schedule in Eq.~\ref{eq:forward}, s.t. $\alpha_T=0$ and $\sigma_t=T$ and switch to velocity prediction \cite{salimans2022progressive, lin2024common}. Together, these changes ensure that the first update step at $T$ is informative and unbiased.

Unfortunately, converting a model to velocity prediction requires extra training efforts. Hence, state-of-the-art few-step methods instead decide to remain in noise prediction mode, but compute loss on $\hat{\mathbf{x}}_0$ \cite{sauer2023adversarial,lin2024sdxl}. While this circumvents the triviality problem of noise prediction at $T$, we argue that it also introduces a bias in the first update step. To see this, consider the first-order update in Eq.~\ref{eq:backward}. The update step $f(\mathbf{x}_t)$ constitutes as a weighted sum of the current estimated signal $\hat{\x}_0$ and the model output $\eps_\student$. Importantly, for noise prediction models, the estimated signal is a function of $\eps_\student$ itself (Eq.~\ref{eq:x0_hat}). Now, since only the former ($\hat{\x}_0$) goes into the loss (see Eq.~\ref{eq:original_grad}) and since there is no signal whatsoever in $\x_T$, the network is explicitly tasked \textit{not} to predict $\eps_\student=\eps$ (which would give an all black image and hence high loss). As a result, using $\eps_\student$ for the second part the update step in Eq.~\ref{eq:backward_iter} biases the denoising process which leads to error accumulations.

To overcome this issue, we present a simple, training-free alternative to switching to zero-SNR velocity prediction~\cite{lin2024common} that allows the usage of noise prediction models without the aforementioned bias. Specifically, by treating $t=T$ as a unique case and replacing $\eps_\student$ with the true noise $\x_T$, the update $f$ is corrected:

\begin{equation}\label{eq:correction}
f_\student(\mathbf{x}_t)=\begin{cases}
          \alpha_{T-1} \hat{\mathbf{x}}_0(\mathbf{x}_T,\hat{\boldsymbol{\epsilon}}_\student,T) + \sigma_{T-1}\boldsymbol{\epsilon} \quad &\text{if} \, t = T, \\
          \alpha_{t-1} \hat{\mathbf{x}}_0(\mathbf{x}_t,\hat{\boldsymbol{\epsilon}}_\student,t) + \sigma_{t-1}\hat{\boldsymbol{\epsilon}}_{\student}(\mathbf{x}_t,t) \quad &\text{if} \, t <  T. \\
     \end{cases}
\end{equation}

We observed that this small modification can significantly improve the estimated colors, resulting in more vibrant and more saturated hues. This effect is particularly pronounced when the number of inference steps is low. We delve further into the effect of noise correction in Sect. \ref{sec:ablations} and provide qualitative comparisons in Figure~\ref{fig:ablation} and Appendix~\ref{app:noise}.

\section{Experiments}
\label{sec:experiments}

To ensure fairness, we use the Emu model~\cite{dai2023emu} as a base for all experiments. Emu is a state-of-the-art model with 2.7B parameters and resolution $768\times768$. We compare our results to previous distillation methods, such as Step Distillation~\cite{meng2023distillation}, LCM~\cite{luo2023latent}, and ADD~\cite{sauer2023adversarial}, by applying them directly on Emu. All models are replacement trained on a commissioned dataset of images.  Since there is no publicly available code for ADD training, we implemented it ourselves based on the details provided in the paper ~\cite{sauer2023adversarial}.

\subsection{Implementation Details}
Like ADD~\cite{sauer2023adversarial}, UFOGen \cite{xu2023ufogen} and Lightning~\cite{lin2024sdxl}, we train our model with an additional adversarial loss for improved image quality. Following ADD, we use the StyleGAN-T discriminator \cite{sauer2023stylegan}. For single step models, we observe better image quality with a U-Net-based discriminator \cite{schonfeld2020u} crafted from the teacher U-Net, in line with UFOGen \cite{xu2023ufogen} and Lightning \cite{lin2024sdxl}.
We choose timesteps $t \in \{999,750,500\}$ and $t \in \{999,666\}$ for our 3-step and 2-step models, respectively. For \lossname, we set $\gamma(t>900) := 990;\; \gamma(900\geq t>500) := 950;$ and $\gamma(t\leq 500):=200$. From there, the teacher model $\teacher$ takes $k=8$ uniformly spaced time steps. Training was conducted for $15$k iterations on $8$ NVIDIA A100 GPUs, using the Adam~\cite{kingma2014adam} optimizer with a learning rate of $5$e$-6$ for the U-Net and $1$e$-4$ for the discriminator. For Emu Baseline model we run DPM++~\cite{lu2022dpmplus} solver with 25 steps if not stated otherwise. 

\subsection{Quantitative Comparison to State of the Art}
\label{sec:quant}
\begin{figure}[t]
\centering\includegraphics[width=\linewidth]{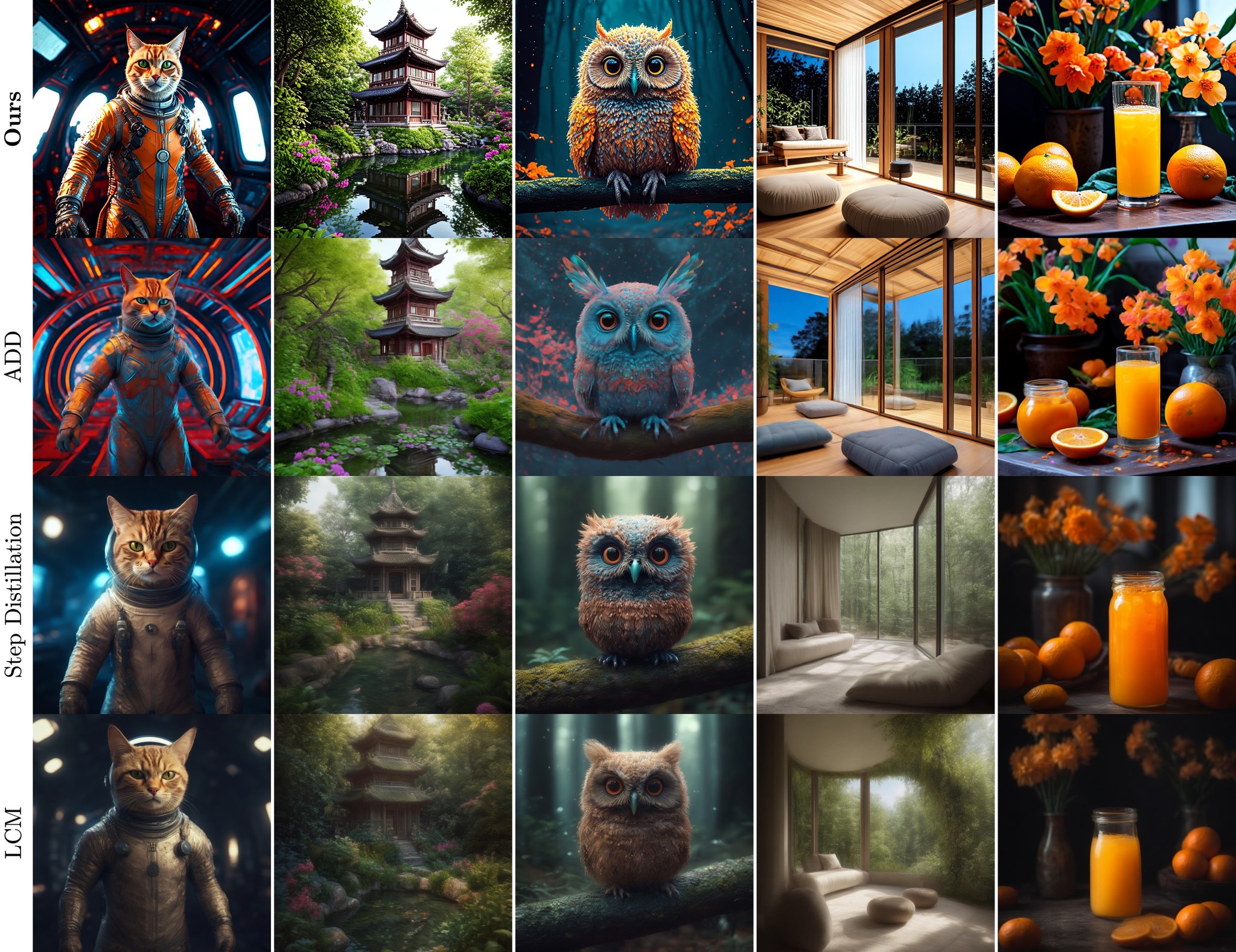}
\vspace{-0.6cm}
  \caption{
   \footnotesize 
   \textbf{\methodname~vs. SOTA}. We show image generations of \methodname~and SOTA methods, all applied to Emu baseline. Every column is generated using the same random seed. \methodname\ provides better realism, sharper images, and a higher level of detail. We attribute these gains to the proposed distillation method, which includes backward distillation, \lossname, and noise correction.
  }
  \vspace{-0.4cm}
  \label{fig:qualitative}
\end{figure}

We compare \methodname\ to previous methods using the FID~\cite{heusel2017gans}, CLIP score~\cite{hessel2021clipscore}, and CompBench~\cite{huang2024t2i}. FID and CLIP measure image quality and prompt alignment, and are evaluated on a split of 5k samples from COCO2017~\cite{lin2014microsoft}, following the evaluation protocoll from~\cite{sauer2023adversarial}. CompBench is a benchmark that separately measures attribute binding (color, shape, and texture) and object relationships (spatial, non-spatial and complex). We generate $2$ images per each prompt in CompBench validation set ($300$ prompts in total). For LCM and \methodname, we compute the  metrics for 1, 2, and 3 steps. For ADD, we compute the metrics for 4 steps, since this method was specifically tuned and configured for 4-step inference, ensuring a fair comparison. We also evaluate Step Distillation for 4 steps to provide a more direct comparison.

Table~\ref{tab:compbench} shows the results. Our 3-step \methodname\ outperforms Step Distillation and ADD in FID, even while using one less step. It also achieves lower FID than LCM for 1, 2, and 3 steps. The CLIP score of our 3-step model is higher than all variants of ADD and LCM and matches the score ($30.2$) of the 4-step Step Distillation model. Unlike Step Distillation and ADD, which degrade FID by $10.1$ and $3.4$ correspondingly compared to the Emu baseline, our 3-step and 2-step \methodname\ preserve FID with slight improvements. For CompBench, any of our 1-, 2-, or 3-step \methodname\ outperforms previous methods in all categories, except color, where 4-step Step Distillation and ADD score similarly to ours. This highlights the superior prompt alignment of \methodname. %

\begin{table}[t!] %

\centering
\caption{
\footnotesize 
\textbf{\methodname~vs. SOTA - Quantitative}. We compare 1-, 2-, and 3-step versions of \methodname\ against the Emu Baseline and other SOTA distillation methods -- Step Distillation, ADD, and LCM, all using the same teacher model and initialization (Emu Baseline). Not only does \methodname\ outperform other methods, but it also excels at preserving baseline FID metric: it remains the same between the baseline and \methodname\ with 3 steps, while for the other methods it decreases significantly.
}
\resizebox{0.9\linewidth}{!}{%
\begin{tabular}{c | c | c c |c c c c c c} 
    \toprule

   \multirow{2}{*}{Method} &\multirow{2}{*}{Steps} &   \multirow{2}{*}{FID  $\downarrow$ } & \multirow{2}{*}{CLIP $\uparrow$ }  & \multicolumn{6}{c}{CompBench  $\uparrow$  (\%)}  \\
    
     &     &    \multicolumn{2}{c|}{ } & \scriptsize{Color}  & \scriptsize{Shape}  & \scriptsize{Texture} & \scriptsize{Complex} & \scriptsize{Spatial} & \scriptsize{Non-spatial} \\ 
    \midrule 
    Emu Baseline & 25 & 35.7 & 30.8 & 51.8 & 39.8 & 53.5  & 46.0   & 60.3  & 67.9  \\ 
    \midrule %
    \midrule
  Step Distill. Emu & 4  & 45.8 & \textbf{30.2} & \textbf{44.1} & 26.9 & 40.0 & 39.6 & 48.7 & 54.3  \\ 
    \midrule
    ADD-Emu & 4 & 39.1 & 29.6 & 43.3 & 32.9 & 44.5 & 36.5 & 42.6 & 56.2  \\ 
    \midrule 
    \multirow{3}{*}{LCM-Emu} & 1 & 123.3 & 24.0  & 29.1& 20.2 & 25.0 & 29.4  & 26.6 & 28.9\\ 
    & 2 & 67.4 & 28.44  &32.6  &	18.9& 25.7 & 33.1 & 39.3 & 43.1 \\ 
    & 3 & 59.1 & 28.94  & 33.8	& 19.0& 25.4& 32.5& 40.8	& 44.2  \\ 
    \midrule 
    
    \multirow{3}{*}{\methodname}  %
    & 1 & 40.4 & 29.0   &  40.9 &  \textbf{37.5} &  46.5 &  41.8 & 53.1 & 59. 5\\ 
    & 2  & \textbf{34.7} & \underline{30.1} &\underline{44.0} &  \underline{36.3} &  \textbf{49.3} &  \underline{42.4} & \textbf{57.9} & \underline{61.4} \\ 
    & 3  & \underline{35.5} & \textbf{30.2} & 42.7 & 36.2 & \underline{47.9} & \textbf{42.8} & \underline{57.4} & \textbf{62.9}  \\ 
    \bottomrule
\end{tabular}
}

\label{tab:compbench}
\end{table}

\subsection{Qualitative Comparison to State of the Art}
\label{sec:qual}

\begin{figure}[t]
\centering\includegraphics[width=\linewidth]{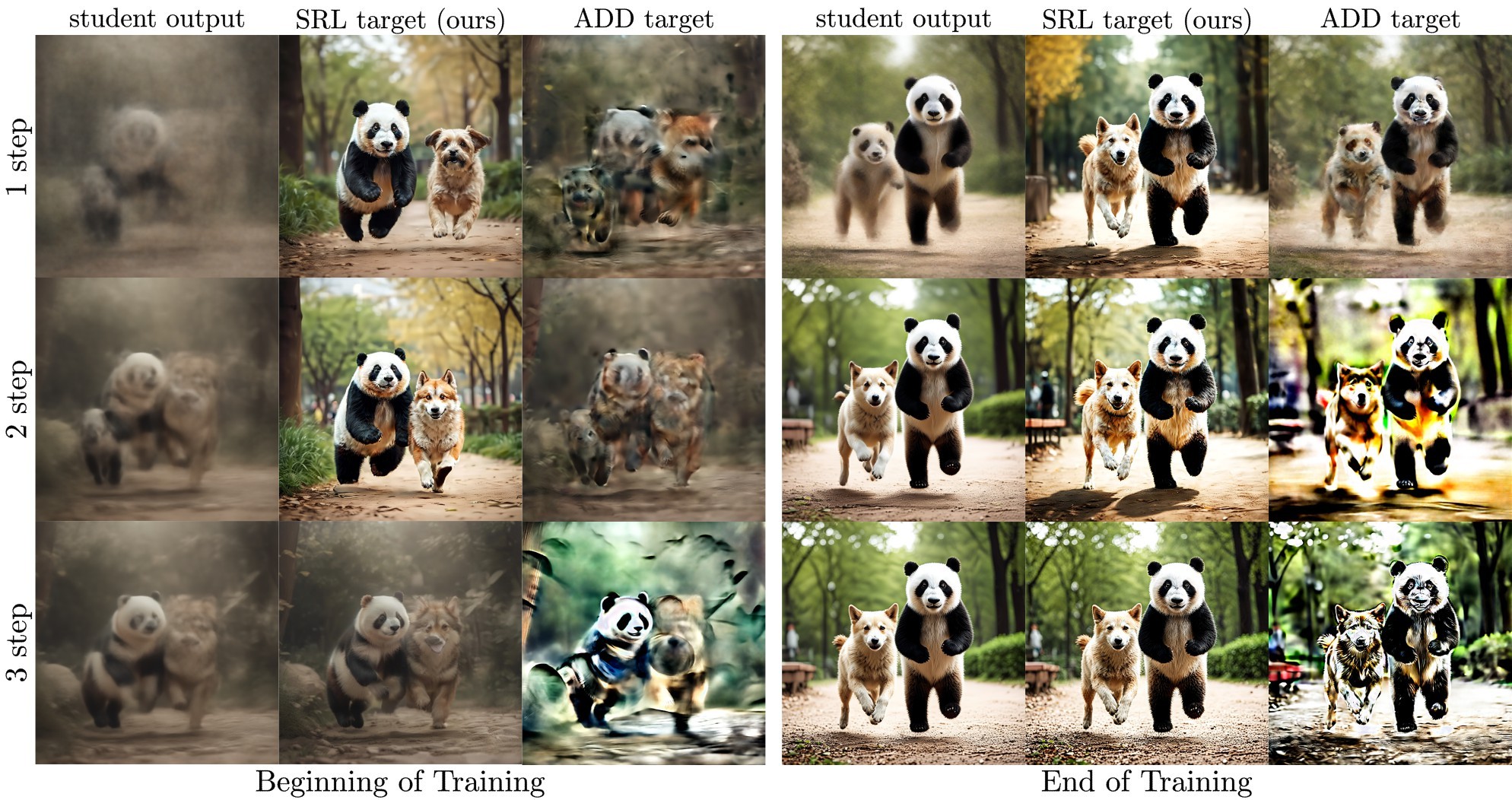}
\vspace{-0.6cm}
  \caption{
   \footnotesize
    \textbf{Qualitative Effect of \lossname}.
   The \lossname\ distillation target consistently encourages the generation of better semantics (step 1), richer colors (step 2), and more high-frequency details (step 3) at all training stages of \methodname. In contrast, we do not observe this effect when using ADD.
   }
   \vspace{-0.1cm}
  \label{fig:loss_qualitative}
\end{figure}

 In Fig.~\ref{fig:qualitative}, we show qualitative comparison of \methodname\ to the current state-of-the-art (SOTA): Step Distillation, LCM, and ADD, all distilling the same baseline Emu model for fair comparison. We observe that ADD images are more crisp than those generated by step distillation and LCM due to the use of an adversarial loss. While both \methodname\ and ADD use a discriminator, \methodname\ produces sharper and more detailed images than ADD.  The enhanced sharpness and detail of \methodname\ are due to our proposed SRL, which effectively refines high-frequency details of the student prediction as depicted in the last row of Fig. \ref{fig:loss_qualitative}.
 
 On the other hand, for ADD, the target images may display a significantly different color spectrum, exhibit color artifacts (see Fig.~\ref{fig:loss_qualitative}), and the colors may fluctuate unpredictably during training iterations. We hypothesize that, to minimize the L2 reconstruction loss in expectation, the ADD model fares best by predicting color values close to zero, leading to pale images and blurry outlines.

 In addition to improving local details, SRL can also correct text-alignment mistakes of the student, as can be seen on the right side of Fig.~\ref{fig:loss_qualitative} (1-step), where the small panda is transformed back into a dog.

\begin{figure}[t!]
\centering\includegraphics[width=\linewidth]{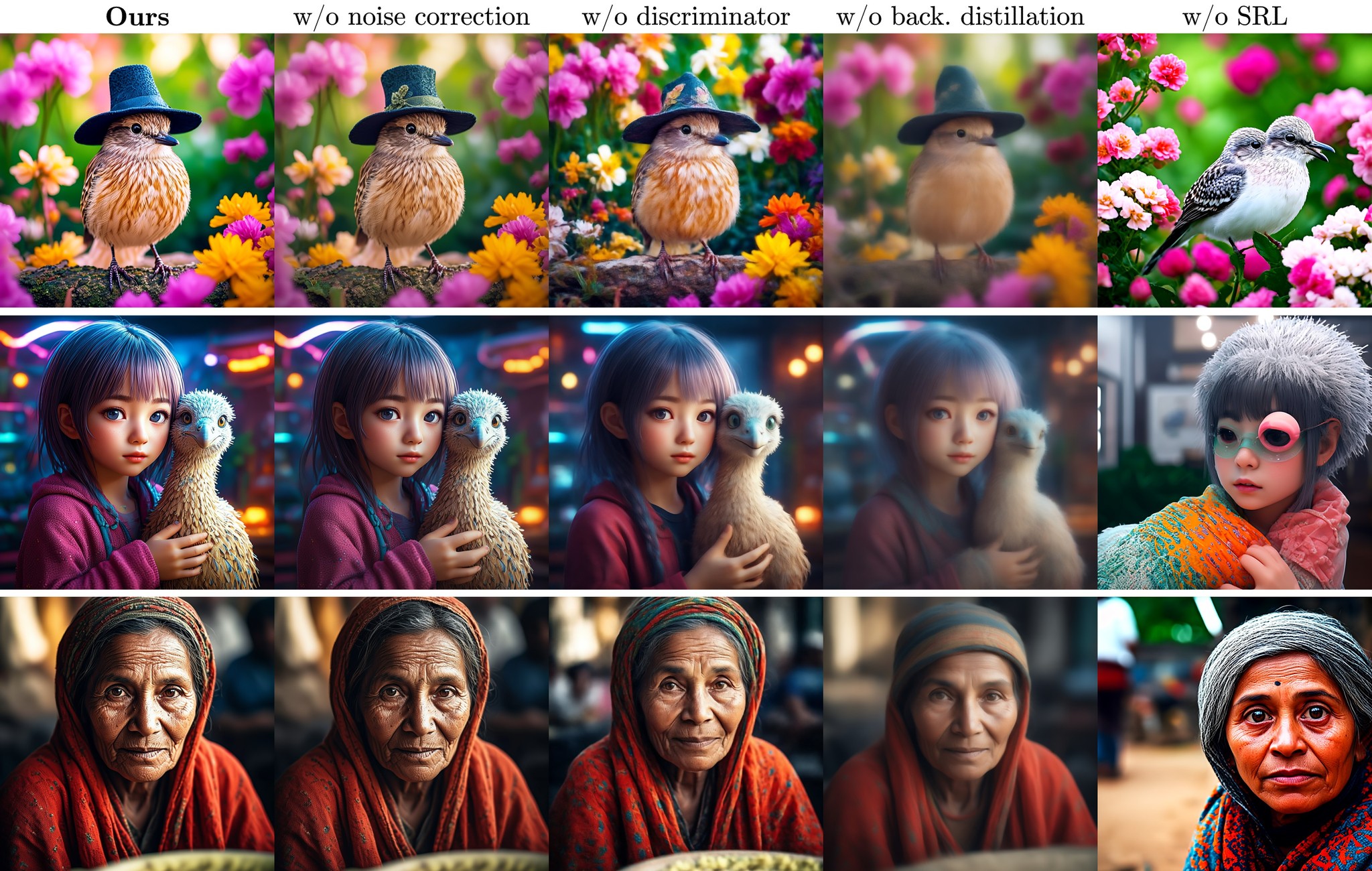}
  \caption{\footnotesize  \textbf{Qualitative Ablations.} We present visual ablation of the components of \methodname. Noise correction improves colors and saturation. The discriminator improves colors and sharpness. However, the biggest impact on sharpness comes from backward distillation. Without the \lossname\ reconstruction loss, the images suffer from artifacts and poor prompt alignment.}
  \label{fig:ablation}
\end{figure}

\subsection{Comparison to Public Models}
We also compare the performance of \methodname~to the public models released by ADD-LDMXL\cite{sauer2023adversarial} and Lightning-LDMXL\cite{lin2024sdxl}.\footnote{This comparison should be interpreted with caution as these methods start from a different base model.
} 
To this end, we compute CLIP and FID scores as detailed in Section \ref{sec:quant} and compare the relative gain/drop versus the base model. Please find the table in Appendix \ref{app:quant}. Our method retains similar text alignment capacity to ADD and Lightning but shows a much more favourable FID increase, especially for two and three steps.

In addition we conducted extensive human evaluations. To this end, we generate images for all methods using three inference steps on $1,000$ randomly sampled prompts from the OUI dataset\cite{dai2023emu}. Paired images are presented to a subset of five from a total of $42$ trained human annotators, who are tasked with voting for the more visually appealing image. The results, aggregated by majority voting, are displayed in Table \ref{tab:human_eval}, where a clear preference for \methodname\ is evident.\looseness=-1

\subsection{Ablations}
\label{sec:ablations}
We conduct quantitative and qualitative ablations on \methodname\ to evaluate the effect of the proposed \textit{backward distillation}, \textit{\lossname}, and \textit{noise correction}. A quantitative evaluation is shown in Table~\ref{tab:main_ablation}, while a complementary visual ablation is provided in Figure~\ref{fig:ablation}.

\textbf{Backward distillation:} Without backward distillation, the CLIP score reduces by two points and FID deteriorates from $35.5$ to $44.2$ (Tab. \ref{tab:main_ablation}), demonstrating the strong positive impact of backward distillation on image quality. As seen in Fig.~\ref{fig:ablation}, this quality degradation is characterized by increased blurriness. \\
\textbf{\lossname:} Without the structure-aware reconstruction loss (i.e. using only the discriminator), FID increases by more than 10 points and CLIP decreases by 4.5 (Tab. \ref{tab:main_ablation}). Fig.~\ref{fig:ablation} shows an increase in artifacts, e.g., a bird with two heads, and worse prompt alignment.

\begingroup

\setlength{\intextsep}{0.0pt plus 0.0pt minus 0.0pt}
\setlength{\tabcolsep}{0.05em} %
\begin{wraptable}{r}{0.55\textwidth}
\vspace{-1.2em}
\caption{\footnotesize \textbf{Human Evaluations} on 1000 randomly sampled prompts from OUI. We report majority voting of 5 human annotators (in \%). The annotators showed a clear preference for our model over ADD-LDMXL \cite{sauer2023adversarial} and Lightning-LDMXL \cite{lin2024sdxl}.}
\resizebox{1.0\linewidth}{!}{%
\begin{tabular}{l c c c } 
\toprule
Method & Win \hspace{2 mm} & Tie \hspace{2 mm} & Loss \\
\midrule 
Ours vs ADD-LDMXL   & \textbf{71.2}  & 1.2 & 27.6 \\ 
Ours vs Lightning-LDMXL \hspace{2 mm} & \textbf{60.6} & 21.0 & 18.4 \\  
\bottomrule
\end{tabular}
}

\label{tab:human_eval}
\vspace{1em}

\caption{
 \footnotesize \textbf{Ablations.} We show the individual effects of every components of \methodname. While noise correction and discriminator have a positive effect on performance, the main gains come from backward distillation and \lossname.
 }

\centering
\small

\begin{tabular}{l c c} 
\toprule
Method & FID $\downarrow$ & CLIP $\uparrow$   \\
\midrule 
Emu Baseline DPM 25 Steps & 35.7 &  30.8  \\ 
Emu Baseline DPM 3 Steps & 66.5 & 25.6  \\  
\midrule 
\textbf{Ours}  & \textbf{35.5} &  \textbf{30.2}  \\  
w/o Noise Correction & \underline{35.5} &  \underline{30.0} \\   
w/o Discriminator & 39.0 &  29.0  \\   
w/o Backward Distillation & 44.2 &  28.2  \\   
w/o \lossname & 45.9 &  25.7  \\   
\bottomrule
\vspace{-6em}
\end{tabular}

\label{tab:main_ablation}
\end{wraptable}

\textbf{Noise correction:} While FID and CLIP are almost not affected by noise correction (Tab. ~\ref{tab:main_ablation}), the effects are clearly seen in the visual ablation in Fig.~\ref{fig:ablation}. Noise correction results in more vivid colors and higher saturation. 

\textbf{Discriminator:} Removing the discriminator slightly blurs the images and reduces color saturation (Fig.~\ref{fig:ablation}). Note that in contrast to previous works, which relay on the same discriminator~\cite{sauer2023adversarial}, in our case the discriminator contributes much less to the image quality thanks to our backward distillation which largely addresses the blurriness problem.

With all components combined, \methodname\ achieves an FID and CLIP close to the original Emu model (see Tab.~\ref{tab:main_ablation}).

\section{Limitations}
\textbf{Constraints of Human Evaluation.}
Our human evaluation is based on a relatively large set of 1000 Open User Input prompts. Each pair of images is shown to a random set of five out of 42 human annotators and results are aggregated by majority voting. However, this approach may not entirely represent the real-world application of the models. The human evaluation of text-to-image models, particularly in terms of aesthetics, is inherently subjective and prone to variability. Consequently, evaluations conducted with a different set of prompts, annotators, or guidelines may yield varying results.

\textbf{General Limitations of Text-to-Image Models.}
Similar to other text-to-image models, our models may occasionally produce biased, misleading, or offensive outputs. We have made substantial efforts to ensure the fairness and safety of our models. These efforts include the construction of balanced datasets, dedicated evaluation for high-risk categories, and extensive red teaming. Despite these measures, potential risks and biases may still exist.

\section{Conclusion}
We presented \methodname, a novel distillation framework enabling high-fidelity few-step image generation with diffusion models. Our approach comprises three key components: Backward Distillation to reduce train-inference discrepancy, a Shifted Reconstruction Loss (SRL) dynamically adapting knowledge transfer per time step, and Noise Correction to enhance initial sample quality.

Through extensive experiments, \methodname\  achieves remarkable results, matching the performance of the pre-trained teacher model using only three denoising steps and consistently surpassing existing methods. This unprecedented sampling efficiency combined with high sample quality and diversity makes our model well-suited for real-time generative applications.

Our work paves the way for ultra-efficient generative modeling. Future directions include extending to other modalities like video and 3D, further reducing the sampling budget, and combining our approach with complementary acceleration techniques. By enabling on-the-fly high-fidelity generation, \methodname\ unlocks new possibilities for real-time creative workflows and interactive media experiences.

\section{Acknowledgement}
This work would not have been possible without a
large group of collaborators who helped with the underlying base model, infrastructure, data, privacy, and the evaluation
framework. We extend our gratitude to the following
people for their contributions (alphabetical order): Eric Alamillo, Andres Alvarado, Giri Anantharaman, Stuart Anderson, Snesha Arumugam, Christine Awad, Chris Bray, Matt Butler, Vijaya Chandra, Li Chen, Lawrence Chen, Anthony Chen, Jessica Cheng, Vincent Cheung, Michael Cohen, Lauren Cohen, Xiaoliang Dai, Andrew Denyes, Abhimanyu Dubey, Cristian Canton Ferrer, Assaf Gelber, Jort Gemmeke, Freddy Gottesman, Freddy Gottesman, Nader Hamekasi, Jack Hanlon, Zecheng He, Zijian He, Tarek Hefny, Ji Hou, Shuming Hu, Jiabo Hu, Ankit Jain, Kinnary Jangla, Abhishek Kadian, Daniel Kreymer, Praveen Krishnan, Carolyn Krol, Kunpeng Li, Tianhe Li, Kevin Chih-Yao Ma, Chih-Yao Ma, Dhruv Mahajan, Dhruv Mahajan, Mo Metanat, Dan Moskowitz, Simran Motwani, Vivek Pai, Mitali Paintal, Guan Pang, Devi Parikh, Hyunbin Park, Vladan Petrovic, Filip Radenovic, Vignesh Ramanathan, Ankit Ramchandani, Stephen Roylance, Kalyan Saladi, Mayank Sanganeria, Dev Satpathy, Alex Schneidman, Shubho Sengupta, Hardik Shah, Shivani Shah, Yaser Sheikh, Mitesh Kumar Singh, Animesh Sinha, Karthik Sivakumar, Yiwen Song, Lauren Spencer, Ragavan Srinivasan, Harihar Subramanyam, Fei Sun, Khushboo Taneja, Kate Trono, Sam Tsai, Mor Tzur, Simon Vandenhende, Jialiang Wang, Mike Wang, Rui Wang, Xiaofang Wang, Mack Ward, Yi Wen, Bichen Wu,Felix Xu, Seiji Yamamoto, Licheng Yu, Matthew Yu, Hector Yuen, Luxin Zhang, Peizhao Zhang, Yinan Zhao, Yue Zhao, Jessica Zhong, and Tali Zvi.

Finally, thank you Manohar Paluri, Ryan Cairns, Connor Hayes and Ahmad Al-Dahle for your support and leadership.

\bibliographystyle{splncs04}
\bibliography{main}
\appendix

\section{Quantitative Results}\label{app:quant}

In Table~\ref{tab:quant} we compare 1, 2 and 3 step versions of \methodname\ against the publically available models of SOTA few-step methods -- LDMXL-Turbo \cite{sauer2023adversarial} and LDMXL-Lightning \cite{lin2024sdxl}. Since our method has a different base model, we report numbers in relative terms to the base model. As can be seen, \methodname\ shows similar CLIP score preservation as Lightning. Turbo's CLIP scores improve over the teacher for the one step model, which is likely to be an artifact of the metric itself rather than a sign of superior text alignment. Regarding image quality, we find that \methodname\  shows significantly less FID score degradation than both Turbo and Lightning. In fact, our 2 and 3 step model show even slight improvements in FID.\\

\begin{table}
\centering

\caption{
\textbf{\methodname~vs. public SOTA - Quantitative}.
}

\begin{tabular}{l c c} 
\toprule
Method & FID $\downarrow$ & CLIP $\uparrow$   \\
\midrule 
LDMXL Baseline \scriptsize{(25 Steps)} & 100$\%$ \scriptsize{(24.50)} &  100$\%$ \scriptsize{(32.01)}
\vspace{0.05cm} %
\\ 
\hdashline 
\noalign{\vspace{0.075cm}}  %
LDMXL-Turbo \scriptsize{(3 Steps)} & 129.9$\%$ \scriptsize{(31.80)} &  99.2$\%$ \scriptsize{(31.76)}\\ 
LDMXL-Turbo \scriptsize{(2 Steps)} & 133.5$\%$ \scriptsize{(32.72)} &  99.8$\%$ \scriptsize{(31.94)}\\
LDMXL-Turbo \scriptsize{(1 Step)} & 132.3$\%$ \scriptsize{(32.40)} &  \textbf{100.5$\%$} \scriptsize{(32.17)}\\
LDMXL-Lightning \scriptsize{(3 Steps)} & 132.8$\%$ \scriptsize{(32.50)} &  98.2$\%$ \scriptsize{(31.44)}\\ 
LDMXL-Lightning \scriptsize{(2 Steps)} & 130.9$\%$ \scriptsize{(32.09)} &  97.1$\%$ \scriptsize{(31.09)}\\ 
LDMXL-Lightning \scriptsize{(1 Step)} & 125.1$\%$ \scriptsize{(30.60)} &  95.2$\%$ \scriptsize{(30.46)}\\ 

\midrule 
\midrule 

EMU Baseline \scriptsize{(25 Steps)} & 100$\%$ \scriptsize{(35.74)} &  100$\%$ \scriptsize{(30.76)}
\vspace{0.05cm} %
\\ 
\hdashline 
\noalign{\vspace{0.075cm}}  %
Imagine Flash \scriptsize{(3 Steps)} & 99.4$\%$ \scriptsize{(35.53)} &  98.2$\%$ \scriptsize{(30.19)}\\ 
Imagine Flash \scriptsize{(2 Steps)} & \textbf{98.1$\%$} \scriptsize{(34.70)} &  98.0$\%$ \scriptsize{(30.14)}\\ 
Imagine Flash \scriptsize{(1 Step)} & 116.3$\%$ \scriptsize{(40.37)} &  94.4$\%$ \scriptsize{(29.03)}\\

\bottomrule
\end{tabular}

\label{tab:quant}
\end{table}

\section{Generation Variance}

In Figures~\ref{fig:bird} and~\ref{fig:panda}, we demonstrate the proficiency of our model in generating a diverse array of high-quality samples, all derived from the same text prompt, with the sole variation being the initial noise.
\begin{figure}[t]
\centering\includegraphics[width=\linewidth]{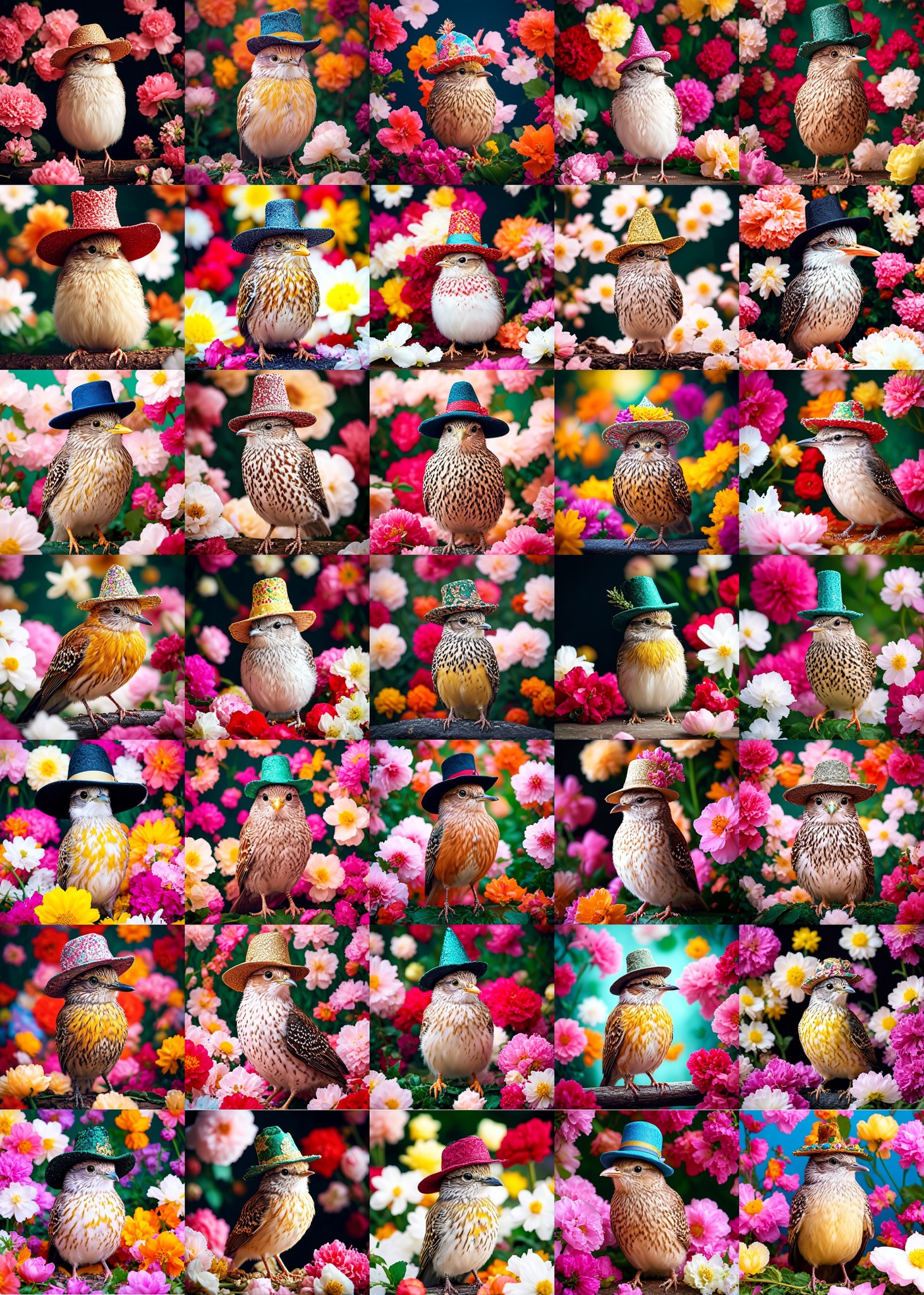}
  \caption{\footnotesize  \textbf{Prompt Variance.} Images generated with the proposed model given the prompt `Cute bird in fancy hat with flowers on the background' with different initial noise.}
  \label{fig:bird}
\end{figure}

\begin{figure}[t]
\centering\includegraphics[width=\linewidth]{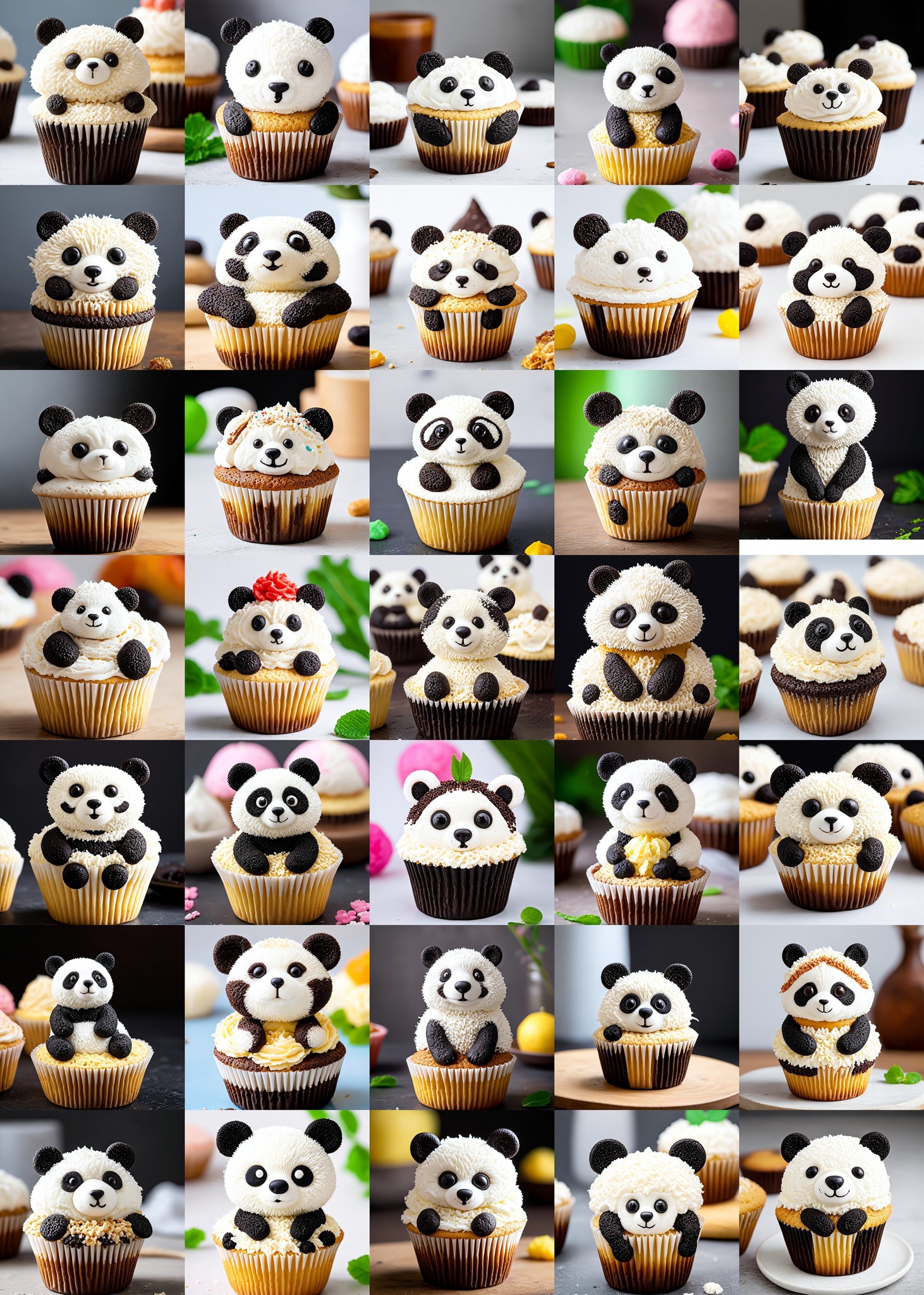}
  \caption{\footnotesize  \textbf{Prompt Variance.} Images generated with the proposed model given the `A cupcake in the shape of a panda' with different initial noises.}
  \label{fig:panda}
\end{figure}

\section{Generated Images}

In Figures~\ref{fig:var1} and~\ref{fig:va2}, we provide further qualitative examples demonstrating the performance of our model.
\begin{figure}[t]
\centering\includegraphics[width=\linewidth]{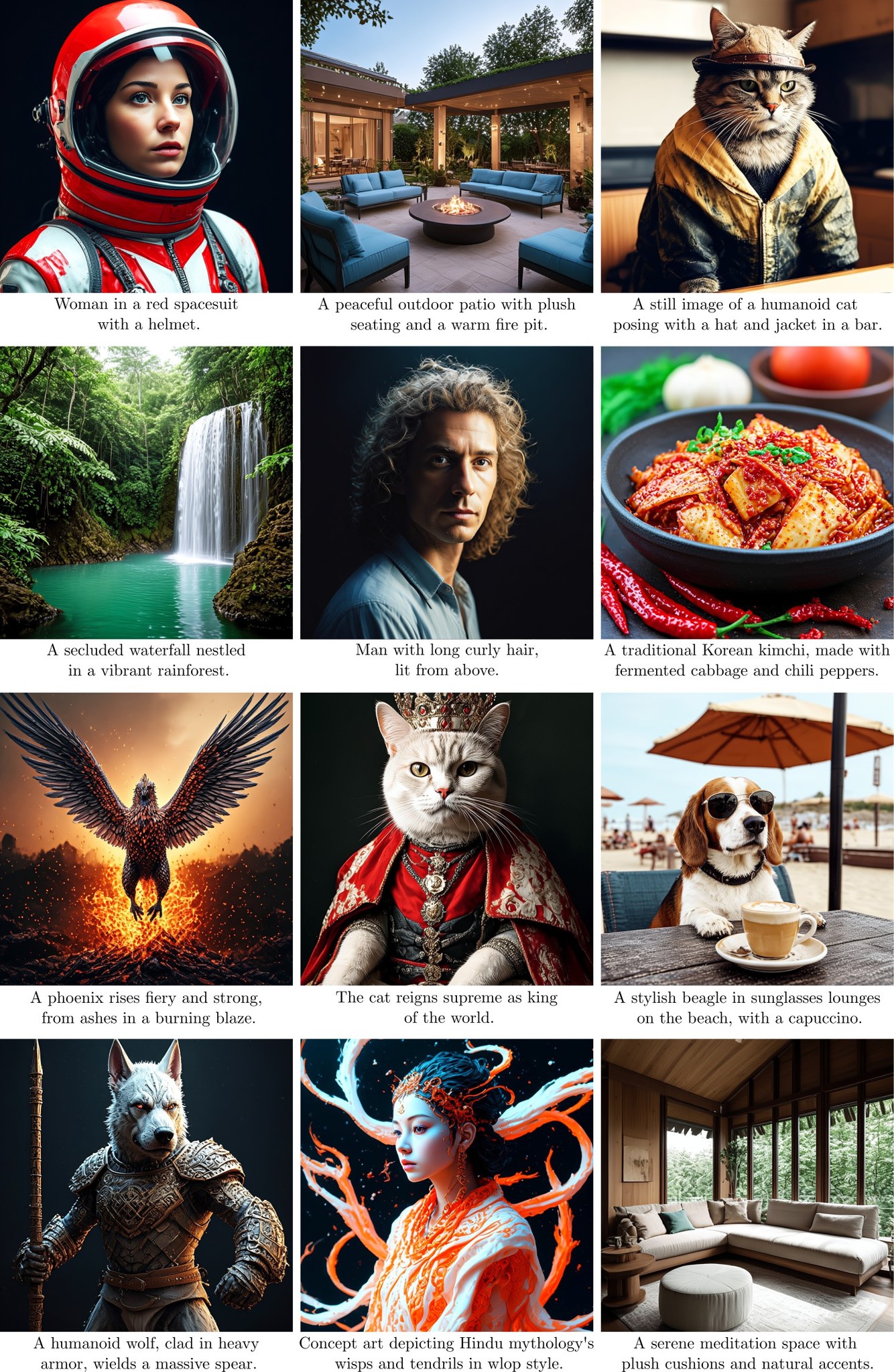}
  \caption{\footnotesize  \textbf{Generated Images.} Images generated with the proposed model. Each with the corresponding text prompt.}
  \label{fig:var1}
\end{figure}

\begin{figure}[t]
\centering\includegraphics[width=\linewidth]{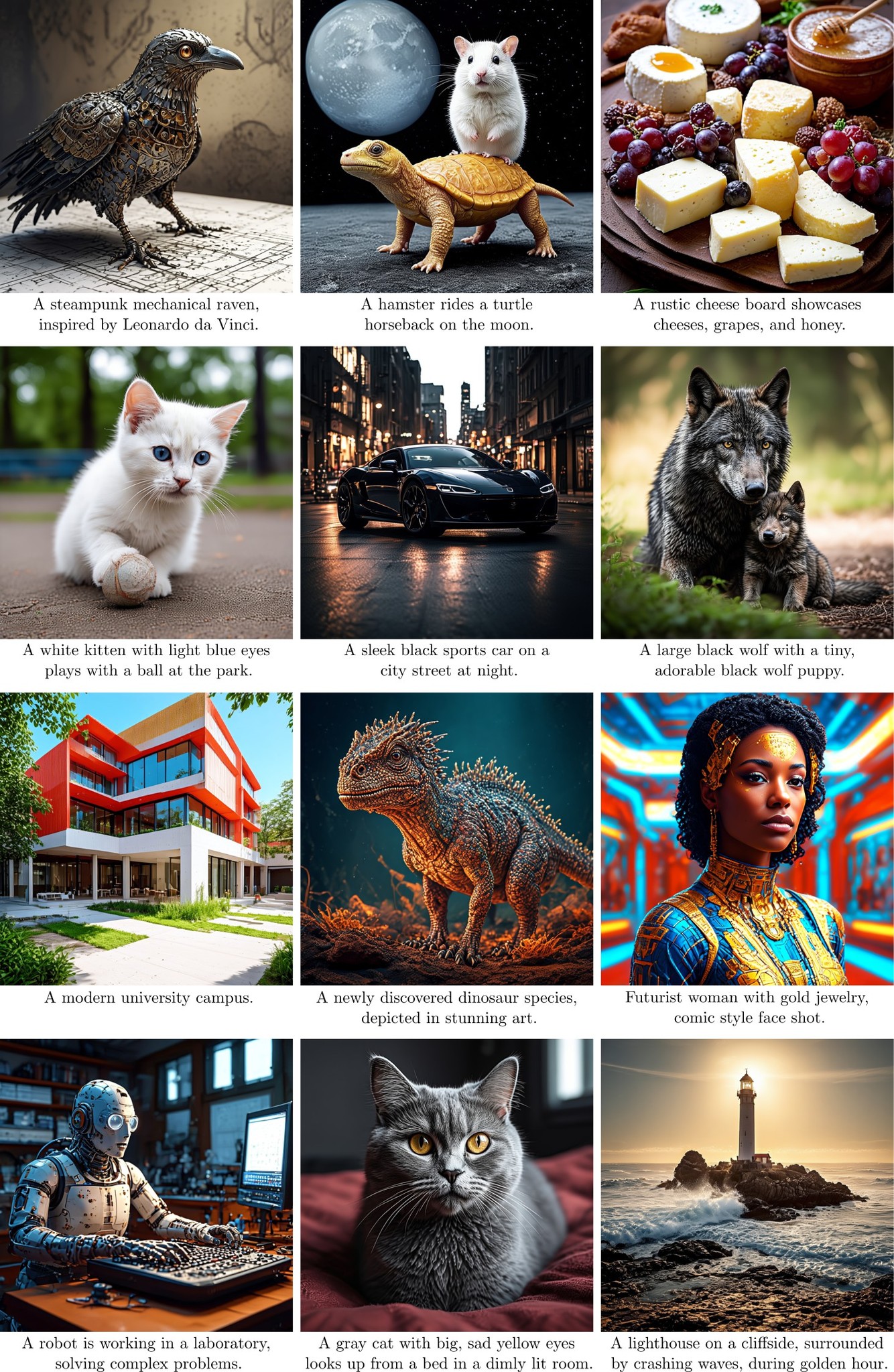}
  \caption{\footnotesize  \textbf{Generated Images.} Images generated with the proposed model. Each with the corresponding text prompt.}
  \label{fig:va2}
\end{figure}

\section{Noise Correction}\label{app:noise}

In Figure~\ref{fig:noise_correction}, we present additional examples illustrating the impact of our proposed noise correction technique. Note the enhancement in image quality, characterized by improved contrast and color saturation. This method further intensifies the dark colors while simultaneously amplifying the brightness of light colors.

\begin{figure}[t]
\centering\includegraphics[width=\linewidth]{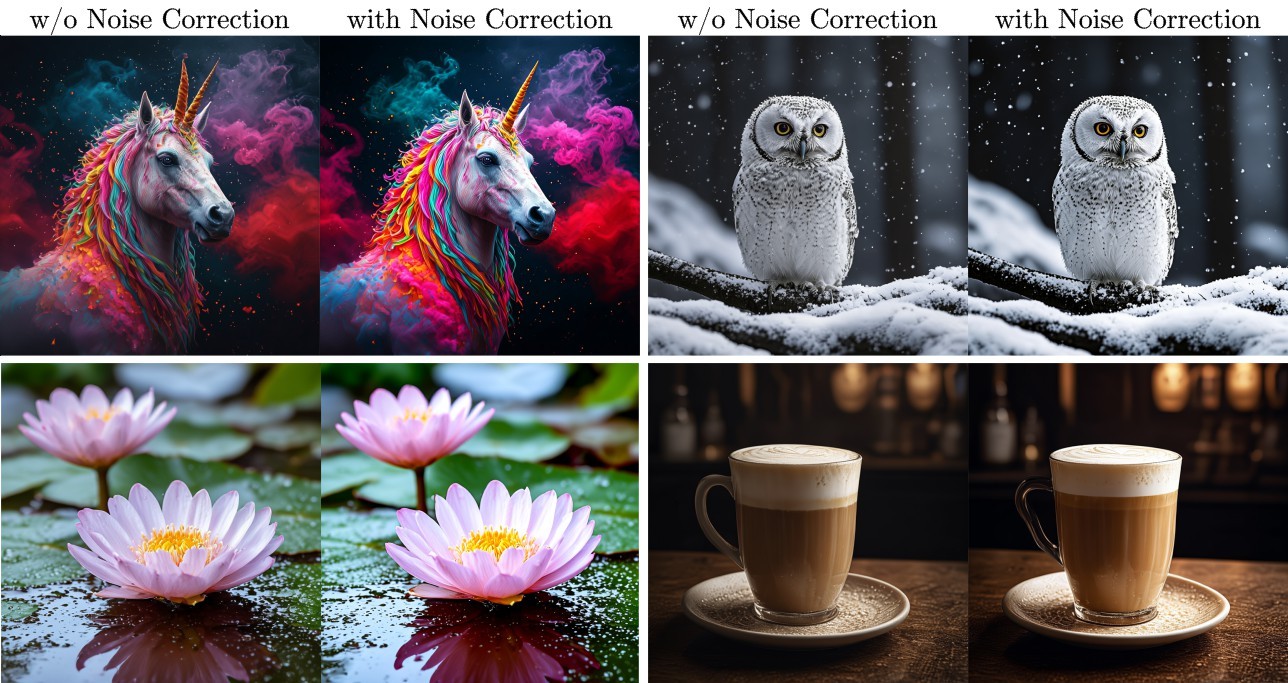}
  \caption{\small Visual demonstration of the effect of Noise Correction. We observe an enhancement in image quality through improved contrast and color saturation. Best viewed on screen.}
  \label{fig:noise_correction}
\end{figure}

\end{document}